\documentclass{article}

     \usepackage[final]{neurips_2020}

\usepackage[utf8]{inputenc} 
\usepackage[T1]{fontenc}    
\usepackage{hyperref}       
\usepackage{url}            
\usepackage{booktabs}       
\usepackage{amsfonts}       
\usepackage{nicefrac}       
\usepackage{microtype}      
\usepackage{graphicx}

\title{Hardware Beyond Backpropagation: a Photonic \\ Co-Processor for Direct Feedback Alignment}

\author{Julien Launay$^{1, 2}$ \hspace{0.4cm} Iacopo Poli$^1$ \hspace{0.4cm} Kilian Müller$^1$ \hspace{0.4cm} \textbf{Gustave Pariente}$^1$ \vspace{0.1cm} \\ \textbf{Igor Carron}$^1$ \hspace{0.4cm} \textbf{Laurent Daudet}$^1$ \hspace{0.4cm} \textbf{Florent Krzakala}$^{1, 2, 3}$ \hspace{0.4cm} \textbf{Sylvain Gigan}$^{1, 4}$
\vspace{0.2cm} \\
 $^1$LightOn \hspace{0.5cm} $^2$LPENS, École Normale Supérieure\hspace{0.5cm} $^3$ IdePhics, EPFL \hspace{0.5cm} $^4$LKB \vspace{0.2cm} \\
  \texttt{\{firstname\}@lighton.ai} \\\vspace{-0.6cm}
}

\begin{document}

\maketitle

\begin{abstract} 

The \emph{scaling hypothesis} motivates the expansion of models past trillions of parameters as a path towards better performance. Recent significant developments, such as GPT-3, have been driven by this conjecture. However, as models scale-up, training them efficiently with backpropagation becomes difficult. Because model, pipeline, and data parallelism distribute parameters and gradients over compute nodes, communication is challenging to orchestrate: this is a bottleneck to further scaling. In this work, we argue that alternative training methods can mitigate these issues, and can inform the design of extreme-scale training hardware. Indeed, using a synaptically asymmetric method with a parallelizable backward pass, such as Direct Feedback Alignement, communication needs are drastically reduced. We present a photonic accelerator for Direct Feedback Alignment, able to compute random projections with trillions of parameters. We demonstrate our system on benchmark tasks, using both fully-connected and graph convolutional networks. Our hardware is the first architecture-agnostic photonic co-processor for training neural networks. This is a significant step towards building scalable hardware, able to go \emph{beyond backpropagation}, and opening new avenues for deep learning. 

\end{abstract}

\section{Introduction}

When strong and scalable priors (e.g. convolutions, attention) are available, increasing model size is enough to significantly increase performance. Recent achievements, like GPT-3 \cite{brown2020language} and GShard \cite{anonymous2021an}, are inline with this so-called \emph{scaling hypothesis}: their better performance is explained by an increased number of parameters, leading to models better suited to large compute budgets. However, as models grow past trillions of parameters, scaling their training is challenging \cite{lepikhin2020gshard}. These models are so large they have to be spread over many compute nodes, using a combination of model, pipeline, and data parallelism. These methods are no free-lunch: they trade between compute, memory, and communication bandwidth, sparking challenges in the extreme scaling of deep learning. These trade-offs are a significant bottleneck, and state-of-the-art approaches such as Megatron-LM \cite{shoeybi2019megatron} are architecture-specific, and still waste most of the available compute resources.

To enable extreme scaling of deep learning, we argue for the exploration of a broader approach, where alternatives to backpropagation are considered. Indeed, alternative training methods can drastically change the aforementioned trade-off landscape. For instance, they may enable local weight updates, leading to an asynchronous backward pass. Furthermore, they can be coupled with the development of \emph{beyond backpropagation} hardware. The proliferation of demand for machine learning products has spawned a resurgence in hardware tailored for training and inference. However, these custom chips are still bound by the constraints of existing training pipelines, and by silicon-based computing limitations. Exploring alternative training methods for extreme-scale deep learning holds the promise of a paradigm shift, both in our approach to scaling models, and in extreme-scale hardware.

\paragraph{Related work} Alternative training methods have been motivated by biological realism (e.g. weight transport problem \cite{grossberg1987competitive}) and by practical considerations (e.g. unlocking the backward pass \cite{jaderberg2017decoupled}). Often, there is an intersection between the two: for instance, synaptically asymmetric approaches with fixed backward weights \cite{lillicrap2016random} enable non-von Neumann computing architectures. We focus here on Direct Feedback Alignment (DFA) \cite{nokland2019training}. Through an asymmetric feedback path, it enables parallelization of weight updates, and places a single random projection at the center of the training process. Communication between layers is unnecessary in the backward pass, and only the final error has to be shared--a smaller quantity to communicate between compute nodes than the full gradients in most architectures. Finally, DFA has been demonstrated on modern deep learning tasks \cite{launay2020direct}. 

To answer the growing needs of large-scale machine learning, a number of inference and training chips have been developed, such as Google's TPU \cite{jouppi2017}. However, these chips are mostly dedicated to genric machine learning workloads. Chips tailored to DFA exist, but are task and architecture-specific, and unable to scale \cite{frenkel202028, han2018, han2019}. All-optical neural networks have been proposed \cite{hughes2018, guo2019}. However, they have only been demonstrated in simulations: among others, the careful tuning of large silicon photonics systolic arrays, or the scalability of optical non-linearities remain open challenges \cite{miller2010optical}.

\paragraph{Contributions} We build the first scalable, beyond backpropagation photonic co-processor based on DFA. We experimentally demonstrate our system on MNIST and Cora, two benchmark tasks for fully-connected and graph convolutional networks, showing our hardware reproduces simulated GPU results. Finally, we highlight future paths towards extreme-scale machine learning with beyond backpropagation hardware. 

\section{Methods}
\paragraph{Direct feedback alignment} In a fully connected network, at layer $i$ out of $N$, neglecting biases, with $\mathbf{W}_i$ its weight matrix, $f_i$ its non-linearity, and $\mathbf{h}_i$ its activations, the forward pass can be written as $\mathbf{a}_i = \mathbf{W}_i \mathbf{h}_{i - 1}, \mathbf{h}_i = f_i(\mathbf{a}_i)$.
$\mathbf{h}_0 = X$ is the input data, and $\mathbf{h}_N = f(\mathbf{a}_N) = \mathbf{\hat{y}}$ are the predictions.  A task-specific cost function $\mathcal{L}(\mathbf{\hat{y}}, \mathbf{y})$ is computed to quantify the quality of the predictions with respect to the targets $\mathbf{y}$. The weight updates are obtained through the chain-rule of derivatives: 
\begin{equation}
    \delta \mathbf{W}_i = - \frac{\partial \mathcal{L}}{\partial \mathbf{W}_i} = - [(\mathbf{W}_{i+1}^{T} \delta \mathbf{a}_{i+1})\odot f'_i(\mathbf{a}_i)] \mathbf{h}_{i-1}^{T}, \delta \mathbf{a}_i = \frac{\partial \mathcal{L}}{\partial \mathbf{a}_i}
\end{equation}

\begin{figure}[b]
    \centering
    \includegraphics[width=0.6\textwidth]{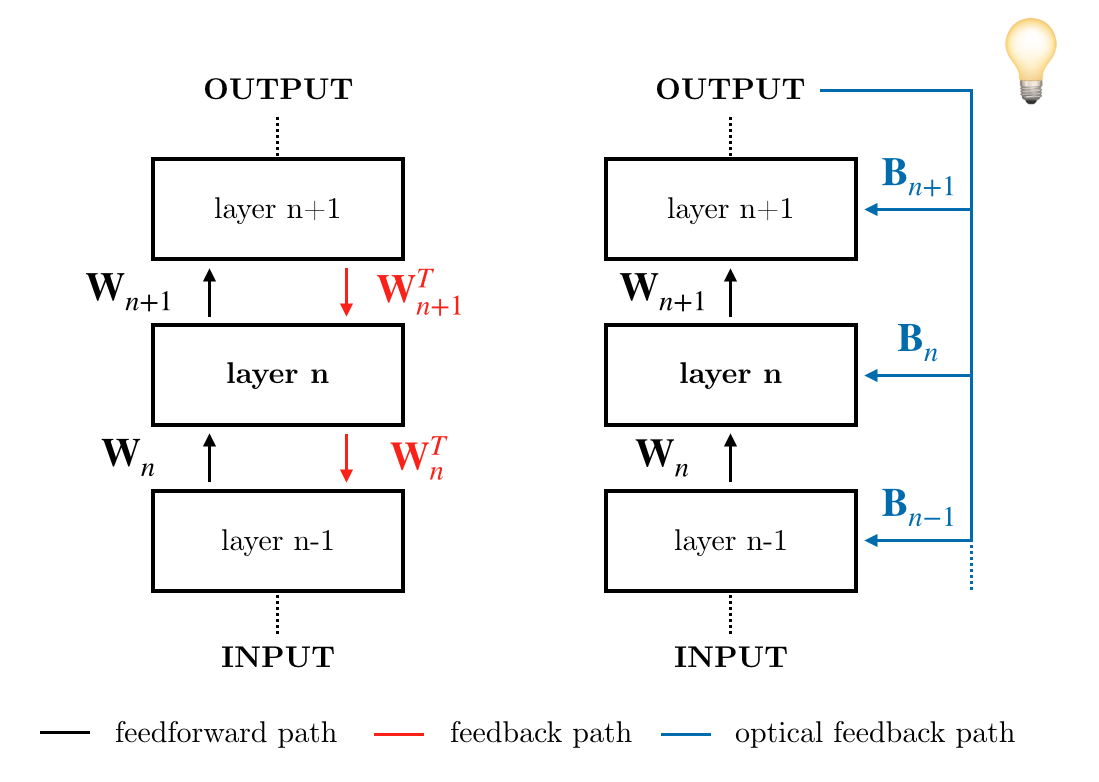}
    \caption{Backpropagation (left) and Optical Direct Feedback Alignment schemes (right). Our co-processor optically implements the random projection $\mathbf{B}\delta \mathbf{a}_y$. The individual feedback for each layer $\mathbf{B}_i\delta \mathbf{a}_y$ can be derived by slicing a single large projection delivered by our device.}
    \label{fig:scheme}
\end{figure}

With DFA, the gradient signal $\mathbf{W}_{i+1}^{T}\delta \mathbf{a}_{i + 1}$ of the (i+1)-th layer is replaced with a random projection of the gradient of the loss at the top layer $\delta \mathbf{a}_y$--which is the error $\mathbf{e} = \mathbf{\hat{y}} - \mathbf{y}$ for common losses: 
\begin{equation}
    \delta \mathbf{W}_i = - [(\mathbf{B}_i\delta \mathbf{a}_y) \odot f'_i(\mathbf{a}_i)] \mathbf{h}_{i-1}^T, \delta \mathbf{a}_y = \frac{\mathbf{\partial \mathcal{L}}}{\partial \mathbf{a}_y}
\end{equation}

\paragraph{Hardware implementation} We perform the random projection $\mathbf{B}\delta \mathbf{a}_y$ optically on our co-processor. The $\delta \mathbf{a}_y$ is collected at the top of the network after the forward pass, and projected using random light scattering. The implemented $\mathbf{B}$ is a Gaussian random matrix, as is commonly used in DFA. Our system is an evolution of the one introduced by \cite{saade2016}, which we modified to use holography to retrieve the full optical field $\mathbf{B}\delta \mathbf{a}_y$ instead of just the intensity $|\mathbf{B}\delta \mathbf{a}_y|^2$. Our co-processor supports an input $\delta \mathbf{a}_y$ and an output $\mathbf{B}\delta \mathbf{a}_y$ with up to 1 and 2 million components respectively. This involves a random matrix $\mathbf{B}$ with trillions of parameters. At this size, it performs the operation in 7 ms--in comparison, a GPU cannot even perform such a large random projection, and a server CPU would take more than a second. At smaller sizes, our system can take down to 1 ms per projection.

A limitation of our optical system is that the physical input $\delta \mathbf{a}_y$ must be binary. While binary DFA has been demonstrated, with the added prospect of forward unlocking in certain cases, it has yet to scale to challenging tasks \cite{frenkel2019learning}. To alleviate this limitation, we ternarize the input to $\{-1;0;1\}$ using a fixed threshold and subtract the projection of the $\{0;1\}$ with the $\{-1;0\}$ vector. While this requires two acquisitions from our co-processor to train once, we also recover a better approximation of the input $\delta \mathbf{a}_y$ angle: for training, the direction information matters the most, not the magnitude.

\section{Experiments} 

\begin{table}[t]
\centering
\caption{Test accuracies on MNIST and Cora for backpropagation (BP), Direct Feedback Alignment (DFA), and shallow training. Our co-processor reproduces the performance of ternarized DFA.}
\label{tab:results}
\begin{tabular}{@{}lccccc@{}}
\toprule
\multicolumn{1}{c}{}    & \textbf{BP} & \multicolumn{3}{l}{\textbf{DFA}}          & \multicolumn{1}{l}{\textbf{Shallow}} \\ \cmidrule(l){2-6} 
\multicolumn{1}{c}{}    &             & vanilla & ternarized & optical ternarized &                                      \\
\textbf{FC-MNIST}       & 98.4            &    97.9     & 98.1           & 97.5                   &  92.4                                    \\
\textbf{GraphConv-Cora} & 82.3        & 80.9    & 81.5       & 80.6              & 48.2                                 \\ \bottomrule
\end{tabular}
\end{table}

\begin{figure}[b]
    \centering
    \includegraphics[width=\textwidth]{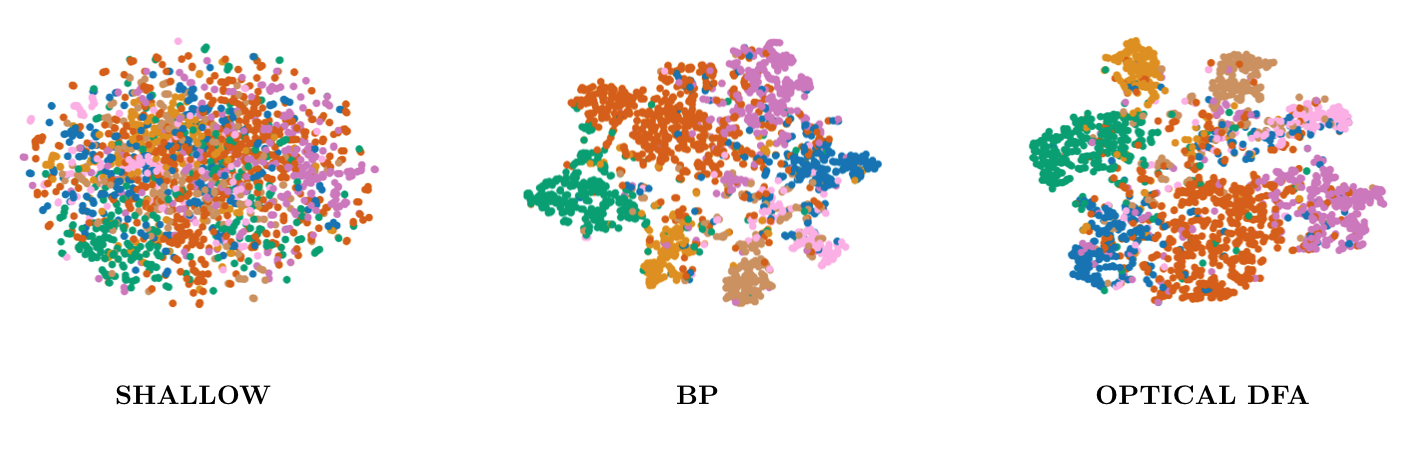}
    \caption{t-SNE visualization of the hidden layer activations of a GraphConv trained on Cora with different methods. Our optical  ternarized DFA builds meaningful embeddings like BP does.}
    \label{fig:tsne}
\end{figure}

\paragraph{Setting} We consider two benchmarks to prove our co-processor delivers random projections appropriate to DFA. The first one consists in a fully-connected 3 layers network, trained on the MNIST handwritten digit recognition task. We do not consider a convolutional variant, as DFA is unable to train them. The second benchmark is a graph convolutional network, using the canonical graph convolution of \cite{kipf2017semi}. We evaluate performance on the Cora citation network classification task \cite{sen2008collective}. In both cases, we also compare to a network where only the topmost layer trained: a \emph{shallow} approach. If DFA is effectively training intermediary layers, it should achieve performance above that threshold. For the graph task, we also visualize intermediary activations with t-SNE \cite{maaten2008visualizing} to verify meaningful representations are learned by our optical implementation of ternarized DFA.

We fine-tune hyperparameters by hand for BP, DFA, and ternarized DFA. For optical ternarized DFA, we keep the hyperparameters of ternarized DFA and only tune the ternarization treshold.

\newpage

\paragraph{Results} Our co-processor achieves results in-line with ternarized DFA on a GPU, showing the random projection it delivers are equally useful (Table \ref{tab:results}). The small performance gap may be explained by the minimal hyperparameter tuning and the analog nature of our system. t-SNE further confirms our results (Figure \ref{fig:tsne}), showing similar cluster separation between the embeddings of the network trained with BP and that trained with ternarized DFA on our photonic co-processor. 

Ternarization comes at no performance cost: on these tasks, ternarized DFA and DFA perform equally well. In more challenging tasks, ternarization could become more penalizing. We foresee hardware improvements allowing higher bitdepth: a faster system would make scaling the quantization to more bits affordable, and a different light modulation technology could allow for multiple bits to be directly encoded. 

Finally, note that the random projections involved for these two benchmarks are small: from 10 to 2048 components for MNIST, and from 10 to 32 components for Cora. Our co-processor is capable of scaling to much larger input and output sizes, up to 1M to 2M components. However, demonstrating optical training at such a large scale involves much more complex architectures, which are beyond the scope of this paper. 

\section{Conclusion and outlooks}
We have demonstrated the first architecture-agnostic photonic co-processor for the training of neural networks. Our co-processor reproduces results obtained on a GPU: it is able to generate linear random projections with light, and to use them to train neural networks with DFA--as demonstrated on two benchmark tasks. We corroborated these results by comparing to a shallow training approach, and by visualizing intermediary representations with t-SNE. While at the small scales considered here our co-processor does not deliver a speed-up over DFA on GPU, our optical approach scales to very large dimensions: up to 1 million inputs and 2 million outputs, beyond the ability of existing hardware. In principle, this is enough to scale to the largest architectures of modern deep learning. For instance, a GPT-3 implementation would require a random projection from size 50,000 to 10,000. 

To enable alternative training methods to further extreme-scale machine learning, significant challenges remain. For one, these methods need to be demonstrated on such complex architectures. The training of Transformers with DFA has started being studied in \cite{launay2020direct}. The authors showed that an hybrid approach, where DFA feedbacks are delivered top encoder/decoder blocks and BP is used within the block, was promising: it still holds the potential for large speed-ups, by enabling entire blocks to be trained in parallel, and comes with an easier-to-manage performance penalty. In the context of extreme-scale machine learning, this kind of approach is ideal: communication within a compute node is fast and affordable; thus, BP can be used. DFA can be used in conjunction with model and pipeline parallelism to prevent communication in-between nodes, at a manageable performance cost. 

Finally, alternative training methods should be considered in conjunction with alternative specialized hardware, able to perform specific operations at much larger scale and higher speed than possible in classic electronics. Hardware based on silicon will be always be bound by the von Neumann bottleneck of memory transfer: because the movement of data (e.g. gradients, weights) requires the charging and discharging of electronic components, strict limits exist on the energy consumption and speed of these systems \cite{miller2017attojoule}. In comparison, the main speed and energy limits of our co-processor are imposed by the modulation and detection bandwidths, and there already exist technology and pathways to significant speed-ups. Considering hardware beyond silicon, based on optics for instance, enables us to go beyond the limitations of common electronics, and envision novel paths for the future of deep learning, outside of the realm of the existing hardware lottery \cite{hooker2020hardware}.

\newpage

\bibliographystyle{unsrt}
\bibliography{opticaltraining}

\end{document}